\newif\ifarxiv   
\arxivtrue      

\ifarxiv
    \documentclass[]{fairmeta}
    \newif\ificlrfinal
    \iclrfinalfalse
    \def\iclrfinalcopy{\iclrfinaltrue}
\else
    \documentclass{article} 
    \usepackage{iclr2024_conference,times}
\fi

\iclrfinalcopy  

\usepackage{amsmath,amsfonts,bm}

\def\eqref#1{equation~\ref{#1}}

\def\1{\bm{1}}

\def\vp{{\bm{p}}}

\def\vx{{\bm{x}}}
\def\vy{{\bm{y}}}

\DeclareMathAlphabet{\mathsfit}{\encodingdefault}{\sfdefault}{m}{sl}
\SetMathAlphabet{\mathsfit}{bold}{\encodingdefault}{\sfdefault}{bx}{n}

\ifarxiv
\else
  \usepackage{url}
  \usepackage[hidelinks]{hyperref}
\fi

\usepackage{graphicx}
\usepackage{booktabs} 
\usepackage{xspace} 
\usepackage{sidecap}

\ifarxiv
\else 
    \usepackage[capitalize,noabbrev]{cleveref}
\fi

\crefname{section}{Sec.}{Secs.}
\Crefname{section}{Section}{Sections}
\crefname{table}{Tab.}{Tabs.}
\Crefname{table}{Table}{Tables}
\crefname{figure}{Fig.}{Figs.}
\Crefname{figure}{Figure}{Figures}
\crefname{appendix}{App.}{Tabs.}

\def\mypar#1{\noindent\textbf{#1.}\hspace{0mm}}

\makeatletter 
\DeclareRobustCommand\onedot{\futurelet\@let@token\@onedot}
\def\@onedot{\ifx\@let@token.\else.\null\fi\xspace}

\def\eg{\emph{e.g}\onedot}

\def\wrt{w.r.t\onedot} 

\makeatother

\title{Better (pseudo-)labels for\\semi-supervised instance segmentation}
 
\ifarxiv
    \author[*]{François Porcher}
    \author{Camille Couprie}
    \author{Marc Szafraniec}
    \author{Jakob Verbeek}

    \affiliation{Meta AI}
    \contribution[*]{Work done while interning at Meta}
    
    \date{\today}
    \correspondence{
 \email{francois\_porcher@berkeley.edu},  
 \email{\{coupriec,jjverbeek\}@meta.com}  
    }

    \metadata[Note]{Appeared at the Practical ML for Low Resource Settings workshop at ICLR 2024.}
    
    \abstract{Despite the  availability of large datasets for tasks like image classification and image-text alignment,  labeled data for more complex  recognition tasks, such as detection and  segmentation, is less abundant.
In particular, for  instance segmentation  annotations are time-consuming to produce, and  the distribution of instances is often highly skewed across classes.
While semi-supervised teacher-student distillation methods show promise in leveraging vast amounts of unlabeled data, they suffer from  miscalibration, resulting in overconfidence in frequently represented classes and underconfidence in rarer ones. Additionally, these methods encounter difficulties in efficiently learning from a limited set of examples. We introduce a dual-strategy to enhance the teacher model's training process, substantially improving the performance on few-shot learning. Secondly, we propose a calibration correction mechanism that that enables the student model to correct the teacher's calibration errors.
Using our approach, we observed marked improvements over a state-of-the-art  supervised baseline performance on the LVIS dataset, with an increase of 2.8\% in average precision (AP) and  10.3\% gain in AP for rare classes. 
}
\else 
    \author{François Porcher\thanks{Work done while interning at Meta.} , Camille Couprie, Marc Szafraniec, Jakob Verbeek 
    \\
    Meta AI
    }
\fi

\begin{document}

\maketitle

\ifarxiv
\else
    \begin{abstract}
    Despite the  availability of large datasets for tasks like image classification and image-text alignment,  labeled data for more complex  recognition tasks, such as detection and  segmentation, is less abundant.
In particular, for  instance segmentation  annotations are time-consuming to produce, and  the distribution of instances is often highly skewed across classes.
While semi-supervised teacher-student distillation methods show promise in leveraging vast amounts of unlabeled data, they suffer from  miscalibration, resulting in overconfidence in frequently represented classes and underconfidence in rarer ones. Additionally, these methods encounter difficulties in efficiently learning from a limited set of examples. We introduce a dual-strategy to enhance the teacher model's training process, substantially improving the performance on few-shot learning. Secondly, we propose a calibration correction mechanism that that enables the student model to correct the teacher's calibration errors.
Using our approach, we observed marked improvements over a state-of-the-art  supervised baseline performance on the LVIS dataset, with an increase of 2.8\% in average precision (AP) and  10.3\% gain in AP for rare classes. 

    \end{abstract}
\fi

\section{Introduction}
Despite large-scale datasets being available for a variety of tasks, such as image classification and image-text alignment, label scarcity is a persistent  issue in ``dense'' recognition tasks, including object detection, semantic segmentation, and  instance segmentation.
Moreover, the distribution of training instances per class is often heavily skewed~\citep{gupta19cvpr}, which results in reduced segmentation performance in the long tail of classes for which there are few training samples.  
To address the lack of training data for the ``long tail'' classes, several directions have been explored, such as resampling or re-weighting of rare classes \citep{chen2023area,lin2017focal}. 
These approaches are inherently limited, though, for rare classes with, say, less than ten training samples.
Others have explored the use of diffusion models to generate additional training instances~\citep{zhao23arxiv}.
The latter, however, relies on the availability of a generative model for rare classes,  without a domain shift \wrt the training data of the segmentation model, which is problematic in itself. 
A third approach is to use semi-supervised learning to  leverage unlabeled data which is often abundantly available.
Although student-teacher distillation approaches have been found very effective~\citep{berrada2024guided,filipiak22arxiv}, the one-hot pseudo-labels provided by the teacher can be sub-optimal due to mis-calibration and biases towards frequent classes.

\begin{SCfigure}[1.3][t]
  \centering
    \includegraphics[width=0.4\textwidth]{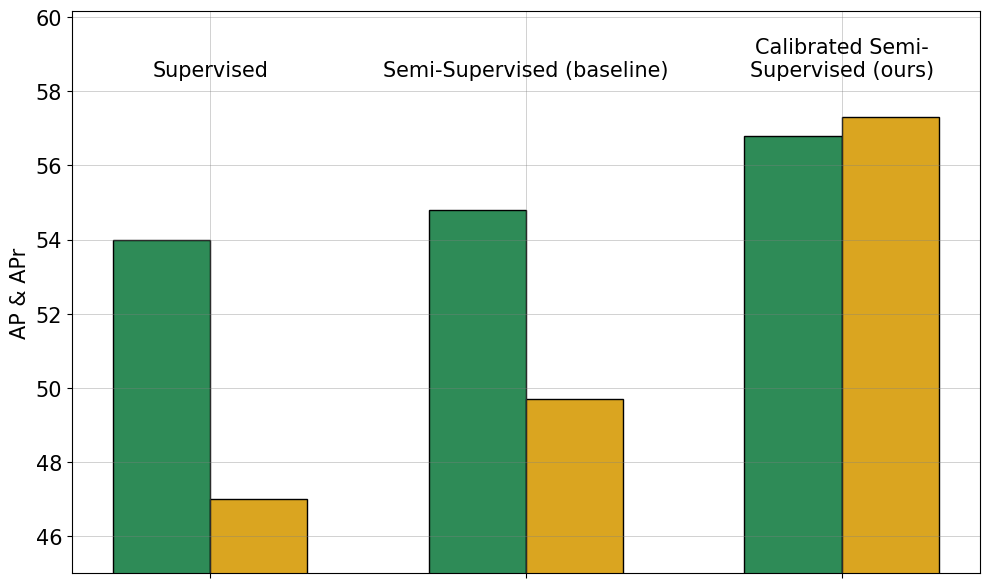}
    \caption{Average precision on LVIS across all classes (green) and rare classes (yellow), for  supervised and  semi-supervised baselines (left, middle), and our approach (right).
    We improve over both baselines, in particular for rare classes where we increase the APr by more than 7 points \wrt the semi-supervised baseline and by more than 10 points over the supervised one.
  }
    \label{fig:teaser}
    \vspace{-2mm}
\end{SCfigure}

In our work we  improve the quality of the pseudo-labels.
First, rather than using one-hot labels, we use label-smoothing  to obtain soft-labels to  train the teacher.
Second, we find that instead of  uniform smoothing it is more effective to  smooth towards similar classes, and  to boost smoothing towards rare classes.
Third, we modulate the soft-labels for the student to explicitly reduce miscalibration of the  confidence scores.
To validate our approach, we conduct instance segmentation experiments on the challenging LVIS dataset, which contains over 1,000 classes, and for more than 300 classes there are less than ten annotated training  images.
Therefore, rather than dropping the labels of part of the training dataset to evaluate our semi-supervised learning approach, we instead augment the LVIS training set with additional unlabeled images. 
Using our approach we obtain marked improvements over semi-supervised learning without our improved distillation labels, and improve the AP of a state-of-the-art supervised baseline performance 10.3 points for rare classes.
While for baselines the AP of rare classes is significantly worse than the overall AP,  our approach  boosts the performance of rare classes to such an extent that this difference disappears, see \cref{fig:teaser}.

\section{Related work}

\mypar{Instance segmentation}
Among class-level recognition tasks, instance segmentation is the most spatially detailed: the goal is to identify each individual  instance and segment it from its background. It generalizes both object detection (which produces instance bounding boxes, but no segmentation) and semantic segmentation (which assigns pixels to classes, but does not separate object instances). 
Most approaches for instance segmentation are derived from object detection methods, by adding a separate branch to the network that for each bounding box produces a segmentation map. This is for example the case for Mask-RCNN~\citep{he17iccv} which extends the Faster RCNN object detector~\citep{ren15nips}.  
Cascade Mask-RCNN~\citep{cai2019cascade}  improves upon Mask-RCNN by applying a series of detector  with increasingly strict IoU detection thresholds.
 Each stage refines the predictions from the previous one, thus improving the accuracy of both bounding box localization and mask generation. The cascade structure is particularly effective for dealing with varying object scales, a common challenge in long-tail datasets.
 In our work we use Cascade Mask-RCNN, with an EVA02 backbone \citep{fang2023eva}, pretrained with   masked-image-modeling, for its superior performance as a basis for our semi-supervised instance segmentation approach.

\mypar{Semi-supervised learning with teacher-student distillation}
In the most basic form of distillation-based semi-supervised learning for instance segmentation, a teacher is pretrained on labeled data only, and then used to  generate pseudo labels for a set of unlabeled images. The student model is then trained on the original labeled data supplemented with the unlabeled images and their pseudo-labels, see \eg \citet{wang22cvpr}. 
To ensure high-quality pseudo-labels  weakly augmented images are fed to the teacher, and  strong augmentations are used for the student to obtain sufficient training signal and generalization. 
Rather than keeping the teacher model fixed during the training of the student, 
it is possible to benefit from the student's improvements to increase the quality of the pseudo-labels. This can be done via  exponential moving average (EMA) updates of the teacher weights from the student model~\citep{filipiak22arxiv,berrada2024guided}.
To mitigate the introduction of noise from lower-quality pseudo-labels,  confidence threshold filtering can be used on the teacher's output. 
Setting the threshold appropriately  is crucial to maintain a good balance between excluding misleading labels and maintaining enough useful training signal on the unlabeled images.
Filtering confidence scores is problematic, however, due to  miscalibration  that results in overconfidence in frequent classes and underconfidence in rare classes. 
This in turn, may result in relatively few pseudo-labels for rare classes, which are precisely the classes for which additional labels are needed the most.
In our work we aim to counter the miscalibration of confidence scores, and so to obtain more useful pseudo-labels for semi-supervised instance segmentation. 

\mypar{Label smoothing} The typical (binary) cross-entropy loss used to train most visual recognition systems can lead to overfitting for big networks.  
Label smoothing  consists in mixing  the one-hot prediction targets with a uniform distribution, which can reduce overconfident, and more generally   miscalibrated, predictions~\citep{szegedy16cvpr,muller2019does}.
Besides uniform label smoothing, \cite{he2022relieving} explored the use of non-uniform label smoothing by mixing one-hot labels with a normalized form of a confusion matrix, that encourages rare classes to be equally predicted, and thus countering bias for frequent classes.
In our approach we also consider non-uniform label smoothing, but rely on similarities between class prototypes in backbone feature space, rather than based on classification scores (of which there are few for rare classes). 
Moreover, we   modulate the (soft) target labels with a term that explicitly counters miscalibration of the confidence scores.

\section{Adaptive label smoothing to set better prediction targets}

Training  with the standard cross-entropy loss with one-hot (pseudo-)labels on a long tail dataset presents several challenges that impede optimal learning and model performance.
First, the training signal is sparse, in particular for rare classes where only few labeled examples are present. 
Second,  pseudo-labels generated by the teacher can be noisy, and a one-hot encoding precludes transmission of information of other  classes that the teacher ranked highly, and which are likely to be  semantically similar (\eg, cats are more likely to confused with  dogs than with, say,  cars).
Finally, a noticeable correlation exists between the frequency of class instances in the training set and the model's confidence scores, leading to overconfidence in frequent classes and underconfidence in rare classes.

To address these issues we smooth the one-hot prediction targets for both teacher and student by explicitly leveraging class similarities and score calibration.
We introduce a \emph{class similarity-based label smoothing} approach for  the teacher model's supervised training phase. 
This technique is used to have the  best starting point possible before using semi-supervised learning.
We develop a \emph{calibration correction} approach for the student distillation phase, which is designed to correct the misscalibration errors made by the teacher model.

\subsection{Leveraging class similarities and frequencies to train a good teacher }

In its most basic form, label smoothing consists in mixing the one-hot prediction target with a uniform distribution~\citep{szegedy16cvpr,muller2019does}. 
This, however, fails to leverage structure in the label space.
We improve upon uniform label smoothing by incorporating class similarity measures. 
By utilizing the model's backbone as a feature extractor, we compute class similarities to enrich the learning target with a more informative signal.
Specifically, let \(\mathcal{B}(\cdot)\)  denote the  backbone network, and \(\mathcal{B}(\vx)\) denote the feature embedding of an object instance  \(\vx\).
We then compute a prototype embedding \(\vp_{i}\) for each class $i$ by averaging the corresponding instance embeddings:
\begin{equation}
    {\vp}_i = \frac{1}{N_{i}} \sum_{n: \vy_n^i=1} \mathcal{B}(\vx_n),
\end{equation}
where \(N_{i}\) represents the  number of instances of class \(i\), and $\vy_n\in\{0,1\}^C$ denotes the one-hot class label associated with the instance  \(\vx_n\), with $C$ the number of classes. 
This average embedding \(\vp_{i}\) effectively encapsulates the collective feature characteristics of class \(i\), thereby facilitating a similarity-based smoothing approach within our model.
We use the prototypes to construct a similarity matrix \(\bf S\), with each elements
$S_{ij} =  \vp_i \cdot \vp_j  / \left( \|\vp_i\| \|\vp_j\| \right) $
given by the cosine similarity between the prototype of class \(i\) and class \(j\).
To  calibrate the confidence scores, in particular for  rare classes, we modulate the similarity scores depending on the number of instance for each class, $N_i$, akin to temperature scaling, before normalizing them with a soft-max:
\begin{equation}
    S'_{ij} = \exp\left(S_{ij}/  N_j^\gamma\right) 
    /
    \sum_{k=1}^{C} \exp\left( S_{ik}  / N_k^\gamma\right),
\end{equation}
where  $\gamma$ controls to what extent the similarities are modulated with the class cardinalities $N_i$ to reinforce smoothing towards rare classes. 
The smoothed labels are then defined as 
\begin{equation}
    \tilde{\vy}   = (1 - \epsilon) \vy   + \epsilon  {\bm S}' \vy,
\end{equation}
 where  \(\epsilon \in [0,1]\) is the mixing weight  between the original one-hot  label and the similarity-based distribution.
In this manner, the label smoothing  incorporates the intrinsic similarities between classes and also adjusts for the disparity in class frequencies, promoting a more balanced learning signal.

\subsection{Calibration-corrected pseudo-labels for better distillation}

Existing semi-supervised distillation approaches often use one-hot labels from the teacher model as pseudo-labels, applying a threshold to filter out low-quality labels~\citep{berrada2024guided,filipiak22arxiv}. 
This, however, overlooks the calibration of confidence scores, which can misrepresent instance quality due to bias towards frequent classes. 
Moreover, naively using soft pseudo-labels from the teacher leads the student model replicating its miscalibration.
To address these issues, we define the class-conditional expected calibration error (CCECE), to capture per class the degree to which a model is overconfident or underconfident, and define it as:
\begin{equation}
\Delta_i = \sum_{b=1}^{B} \frac{N_{i,b}}{N_i} ( \text{acc}_{i,b} - \text{conf}_{i,b} ),
\end{equation}
where \(B\) represents the number of bins in which the $[0,1]$ confidence interval has been split, \(N_{i,b}\) is the number of samples in bin \(b\) for class \(i\), as above \(N_i\) is the total number of samples for class \(i\), \(\text{acc}_{i,b}\) is the accuracy within bin \(b\) for class \(i\), and \(\text{conf}_{i,b}\) is the average confidence in bin \(b\) for class \(i\). 
A negative CCECE for a given class means that the model tends to be overconfident, and vice-versa.
Building on the concept of Expected Calibration Error (ECE) introduced by \cite{naeini2015bayesianbinning}, our CCECE offers a nuanced perspective by assessing calibration errors for each class individually, rather than in aggregate. Unlike ECE, CCECE does not apply an absolute value to the difference between accuracy and confidence, thereby revealing the direction of miscalibration. 
We use the CCECE to adjust the soft pseudo-labels produced by the teacher, $\vy_\textrm{teacher}$, as 
\begin{equation}
\tilde{\vy}_\textrm{teacher} = \vy_\textrm{teacher} + \lambda {\bm\Delta},
\end{equation}
where $\bm\Delta$ the the vector that concatenates all the per-class CCECE values $\Delta_i$, and 
 \(\lambda\) is a tuning parameter that controls the extent of calibration correction applied. 
By leveraging CCECE, we  can improve the calibration of the student model, thereby enhancing the knowledge distillation process between the teacher and student models.
\section{Experimental evaluation}

\subsection{Experimental setup}

\mypar{Dataset}
We conduct experiments on the LVIS v1.0 dataset~\citep{gupta19cvpr}, which contains instance-level segmentations of 1,203 categories.
It has many rare classes, making it ideal for evaluation of models in long-tail distribution scenarios.
For semi-supervised training experiments, we complement LVIS with unlabeled images. 
In particular, we compute  DINOv2 features~\citep{oquab2023dinov2} for object crops of rare classes in LVIS, and use these as queries  to retrieve  64 neighbors with the  Faiss  library~\citep{douze2024faiss} in a large internal dataset.
After removing duplicates, this results in  approximately 200k unlabeled images which we add to the LVIS training set.

\mypar{Evaluation protocol}
We report results on the LVIS validation set, as evaluation on the  LVIS (withheld) test set labels is not longer supported.
To avoid overfiting the validation set, we train our models on 85\% of the official train split, use the remaining 15\% for validation, and report results on the official validation set. 
We report the standard instance segmentation AP metric, and also include the APr metric which evaluates the AP for ``rare'' classes with ten or less instances in the train set.

\mypar{Architecture and training}
We use Cascade Mask RCNN~\citep{cai2019cascade} as instance  segmentation model with an EVA-02 backbone~\citep{fang2023eva}.
To train the teacher model, we tune the class frequency scaling parameter $\gamma$ and the  teacher label smoothing parameter  $\epsilon$ using the 15\% of LVIS train that we have left out, and then retrain the model on the full training set with these parameters. 
Once the teacher is pretrained, we similarly tune the student calibration parameter $\lambda$, and the relative weight of the loss terms for supervised and unsupervised images for the student.
Using a small grid search leads us to set  $\epsilon = 0.1, \gamma = 1.5, \lambda = 2$, and equal weighting between the loss for supervised and unsupervised images.
The teacher is trained on two nodes of eight V100 GPUs each, and takes approximately 10 hours. 
During the semi-supervised learning step, we train the teacher-student ensemble on one node with eight V100s, which takes approximately 24 hours. We use a cosine learning rate schedule, with an initial rate of 4e-5, and a batch size of two per GPU.

\begin{SCfigure}[1.3][t]
\centering
\includegraphics[width=0.37\linewidth]{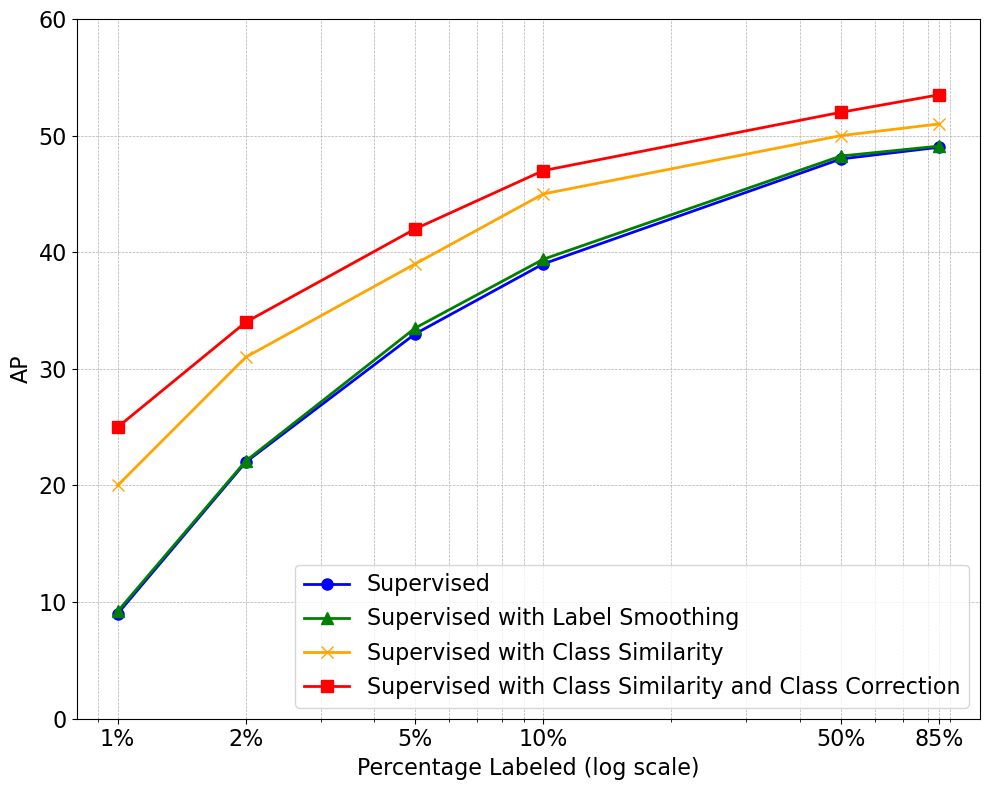} 
\includegraphics[width=0.37\linewidth]{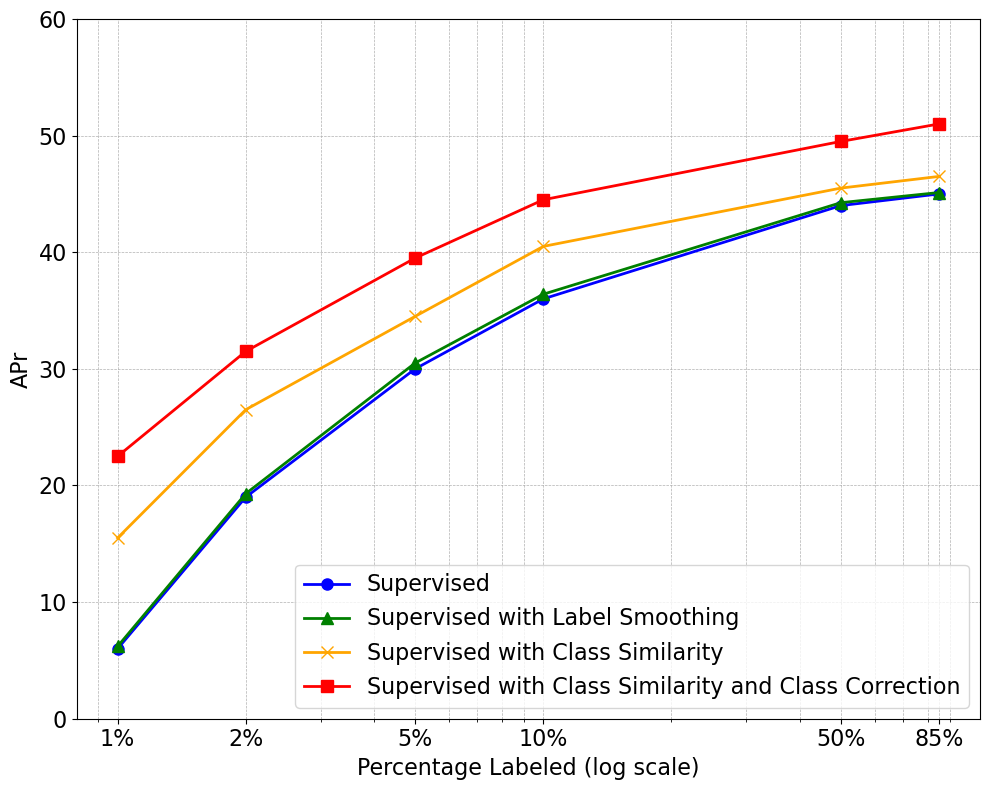}
\caption{Performance of  teacher pretraining on supervised data only in terms of  AP (left)  and APr (right) as a function of the percentage of  LVIS training set used.
}
\label{fig:supervised_ablation}  
\end{SCfigure}

\subsection{Experimental results}

We  demonstrate the influence of our class similarity-based label smoothing  and class correction for the supervised  pretraining of the  teacher in \cref{fig:supervised_ablation}. 
We train models  using different percentages of the  labeled data  to study robustness in low annotation settings. 
While uniform label smoothing does not improve model performance, our class similarity-based label smoothing results in a large boost, up to 10 points in AP and APr when using 1\% of annotations. 
Adding the class correction term further improves results, in particular for rare classes where it raises the APr by about 5 points.

In our  ablations in \cref{tab:student_performance} we observe  boosts of +0.8 AP and +2.7 APr for our baseline semi-supervised model compared to our best supervised teacher model. 
When we use soft labels for distillation, the AP further improves by 0.7 points and APr by 3.3 points, showing the importance of applying soft targets in both models.
Finally, adding the class-conditional  calibration correction boost overall AP by another 1.3 points, and by 4.3 points 
for rare classes bringing it to 57.3 which is a level comparable, and even slightly better,  than the overall AP of 56.8 for this model.   

Our results improve over state-of-the-art results also reported in \cref{tab:student_performance}.
Note that unlike the backbone used by \cite{fang2023eva}, ours was not pretrained using external detection or segmentation annotations, so that it can be directly compared to other models trained on LVIS only.

\begin{SCtable}[1.3][t]
\centering
\small
\begin{tabular}{lrr}
\toprule
                                           Model &   AP &  APr \\
\midrule
EVA, extra data \citep{fang2023eva}  & 55.0 & 52.5 \\
CO-DETR \citep{zong2023detrs} & 56.0 & 53.1 \\
\midrule
Supervised + non-unif.\ smooth.\ + class corr.& 54.0 &     47.0 \\
                Semi-supervised baseline, one-hot labels  & 54.8 &     49.7 \\
               + soft student targets & 55.5 &     53.0 \\
+ soft student target + calibration correction & {\bf 56.8} & {\bf 57.3} \\
\bottomrule
\end{tabular}
\caption{Comparisons to state-of-the-art and ablations. 
EVA \citep{fang2023eva} uses a backbone trained using dense annotations from COCO and Objects365, and AP was computed using the top 1,000 instances. 
Our  backbone was not pretrained on dense annotations, and we use   300 instances for AP computation.}
\label{tab:student_performance}
\end{SCtable}

\begin{SCfigure}[1.3][th]
\centering
  \includegraphics[width=.35\linewidth]{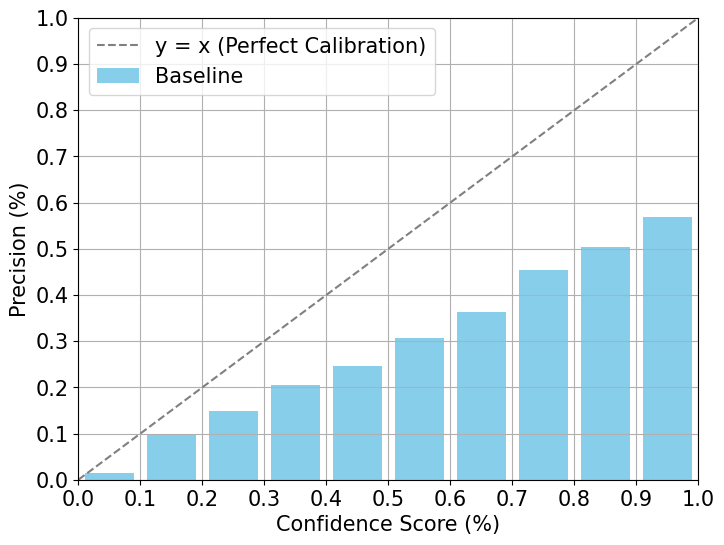}
  \includegraphics[width=.35\linewidth]{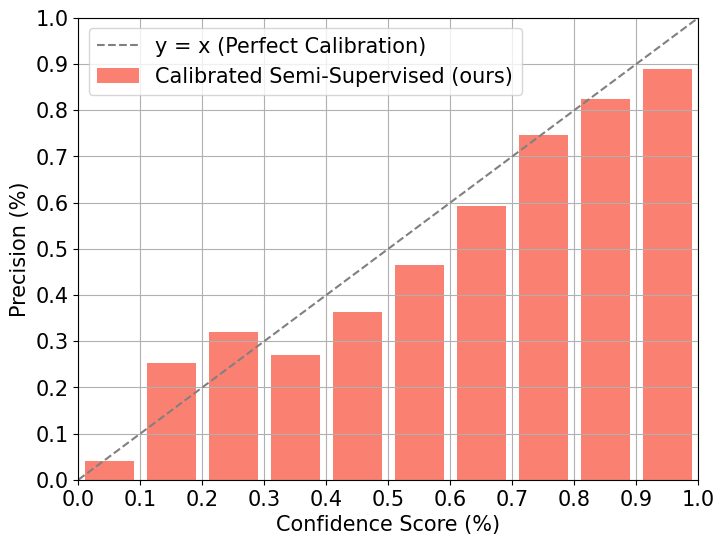}
  \caption{Calibration of our  supervised teacher (left)  and semi-supervised student (right).
  While the teacher model  is  consistently overconfident,  the calibration correction in the distillation alleviates this for the student.
  }
  \label{fig:calibration_ours}  
\end{SCfigure}

In \cref{fig:calibration_ours} we analyse   calibration of our   teacher model (with class similarity smoothing and correction)  and our semi-supervised model with calibration correction.
The  plots show that the systematic overconfidence observed in  the teacher model is by and large corrected  during distillation.
This is also reflected in the aggregate ECE metric which is  0.22 for the teacher  and  0.07 for the student.

\section{Conclusion}

We present two successful strategies to improve the performance of instance segmentation in few-shot learning. The first strategy is to enhance the teacher model's pre-training using class-similarity smoothing,  amplification of smoothing towards  rare class, and a targeted selection of unlabeled images through similarity searches. The second strategy introduces a novel calibration correction mechanism, which enables the student model to rectify calibration errors from the teacher model.
The resulting approach boosts the performance of rare classes,
with ten image annotations or less, to match the average performance across all classes on LVIS.

\ifarxiv
    \bibliographystyle{assets/plainnat}
\else
    \bibliographystyle{iclr2024_conference}
\fi

\ificlrfinal
\subsubsection*{Acknowledgments}
We would like to thank Tariq Berrada for his help on this project.
\fi

\bibliography{biblio}

\ificlrfinal
    \appendix

\section*{Appendix}
\label{app:retrieval}

EVA open sources different backbones. In this work, we build on the ViT-L backbone  \href{https://huggingface.co/Yuxin-CV/EVA-02/blob/main/eva02/pt/eva02_L_pt_m38m_p14to16.pt}{\texttt{eva02\_L\_pt\_m38m\_p14to16}} trained only of masked image modeling on the ImageNet-21K, CC12M, CC3M, Object365, COCO, ADE datasets, without using any bounding-box or segmentation annotations.
We also use this backbone to compute the prototypes embeddings.

In \cref{fig:scatter}, we show the impact of our corrected semi-supervised approach on the accuracy average per class. We note the strong AP improvement on the rare classes.    

In \cref{fig:original}, we display some qualitative examples or our results compared to the supervised pre-trained teacher model. Rare object such as ``pitcher'', ``scissors'', ``traffic light'', and ``curtains'' that were missed by the teacher, are  well segmented with our semi-supervised student model.  

\begin{figure}[ht]
\centering
\begin{tabular}{cc}
\includegraphics[width=0.48\linewidth]{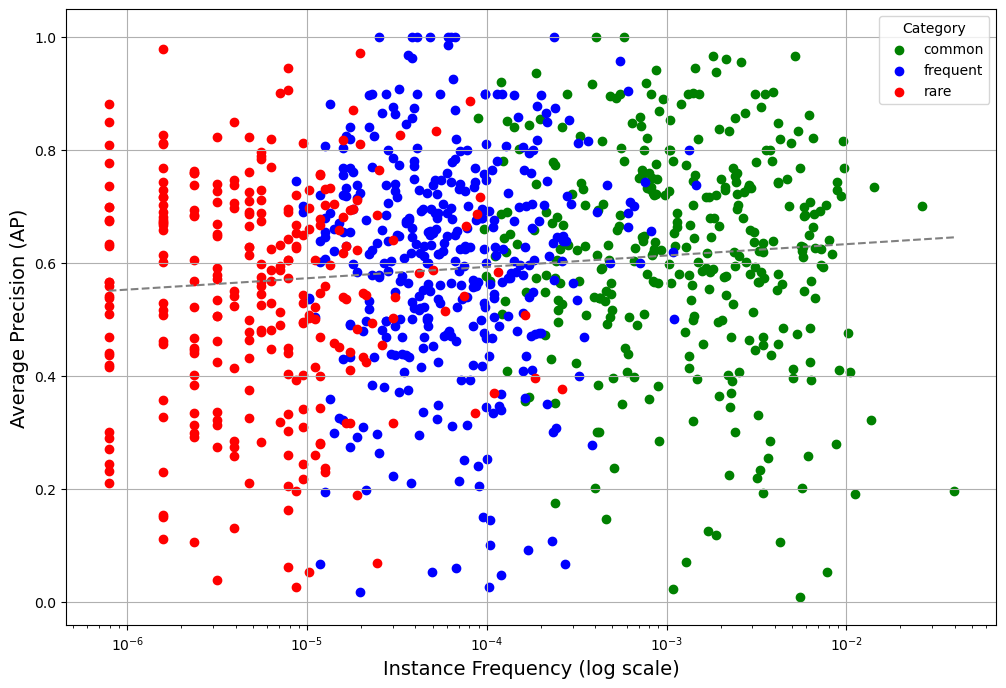}&  
\includegraphics[width=0.48\linewidth]{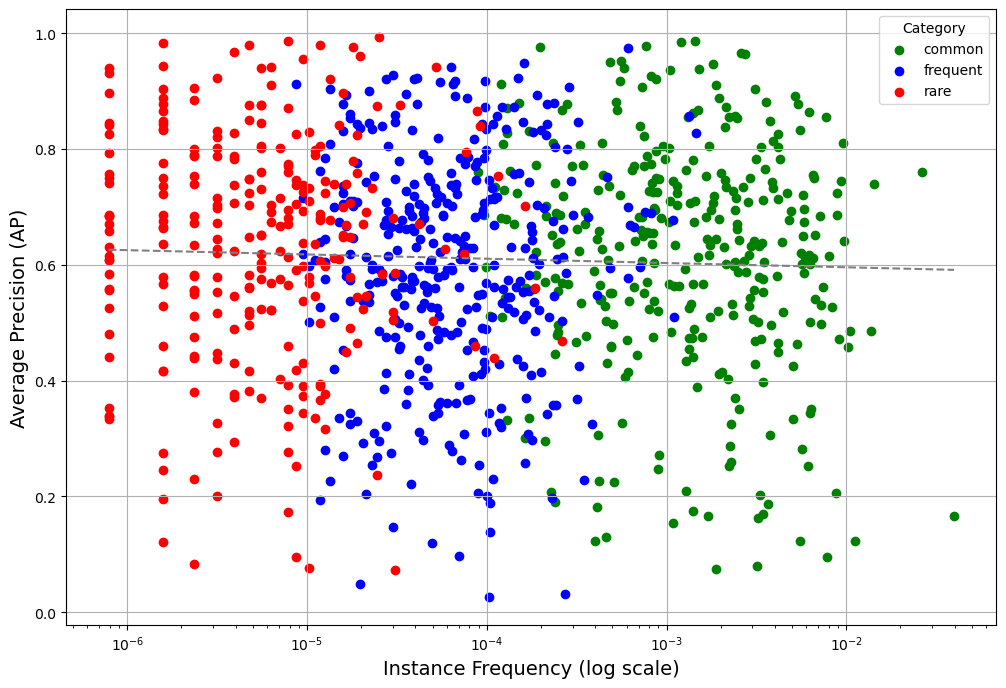}\\
Our supervised pre-trained teacher model & Our semi-supervised student model  \\
\end{tabular}
\caption{Scatter plot of AP of rare, common and frequent classes as function of instance frequency.}
\label{fig:scatter}  
\end{figure}

\begin{figure}
\centering
\begin{tabular}{ccc}
\includegraphics[width=0.32\linewidth]{input_rank17_idx17.jpg}&
\includegraphics[width=0.32\linewidth]{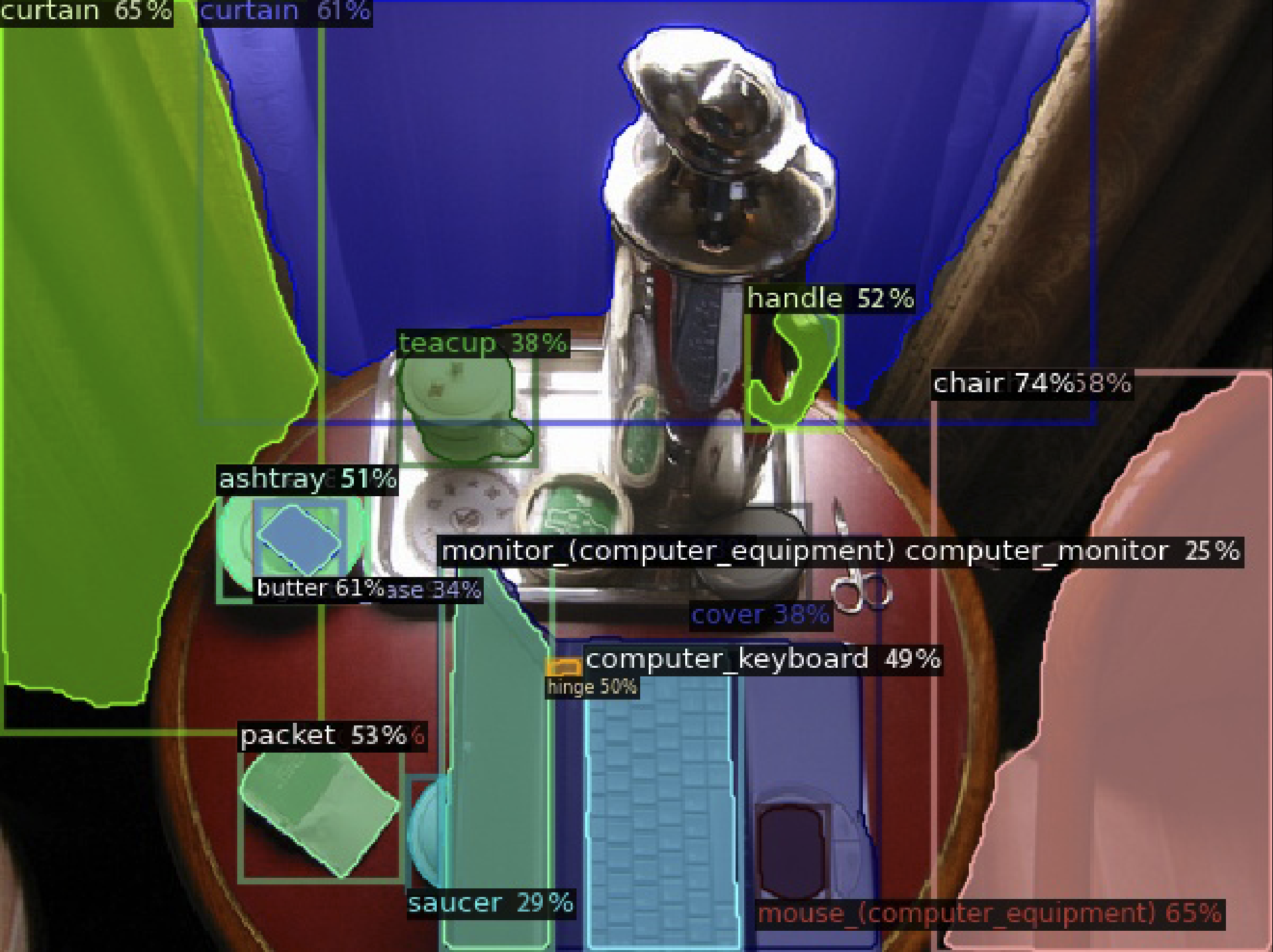}&
\includegraphics[width=0.32\linewidth]{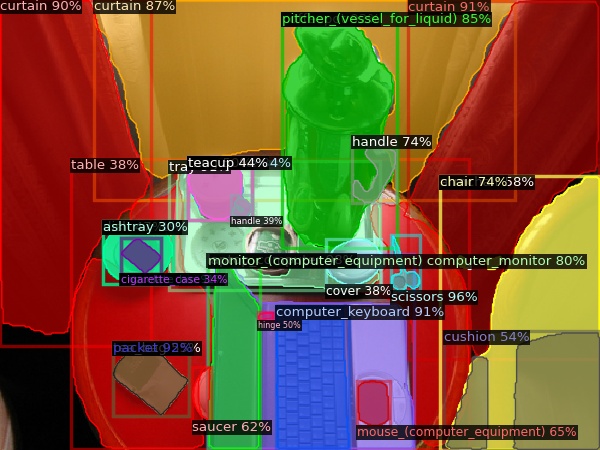}\\
\includegraphics[width=0.32\linewidth]{input_rank19_idx19.jpg}&
\includegraphics[width=0.32\linewidth]{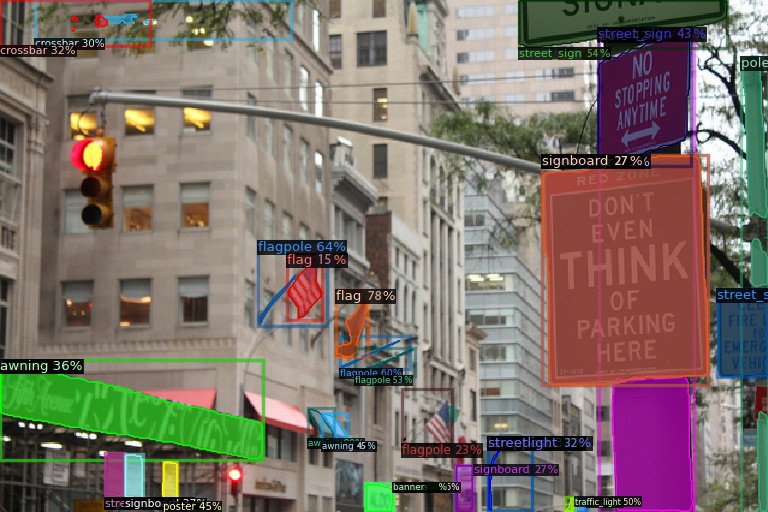}&
\includegraphics[width=0.32\linewidth]{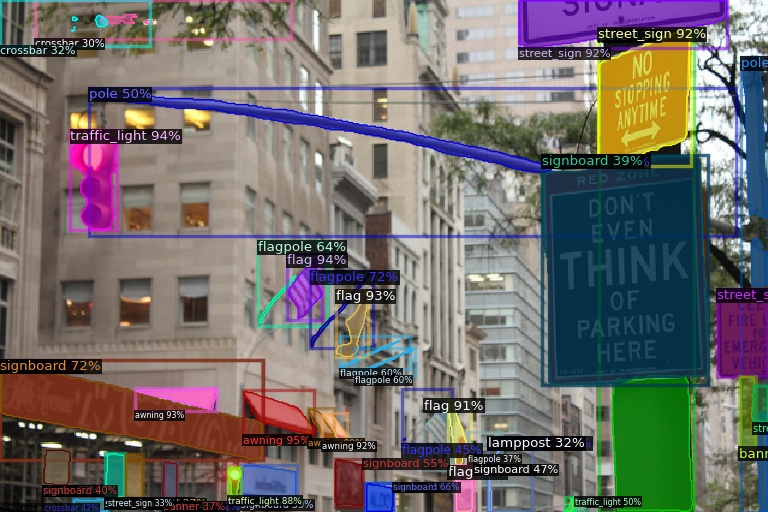}\\
\includegraphics[width=0.32\linewidth]{input_rank21_idx21.jpg}&
\includegraphics[width=0.32\linewidth]{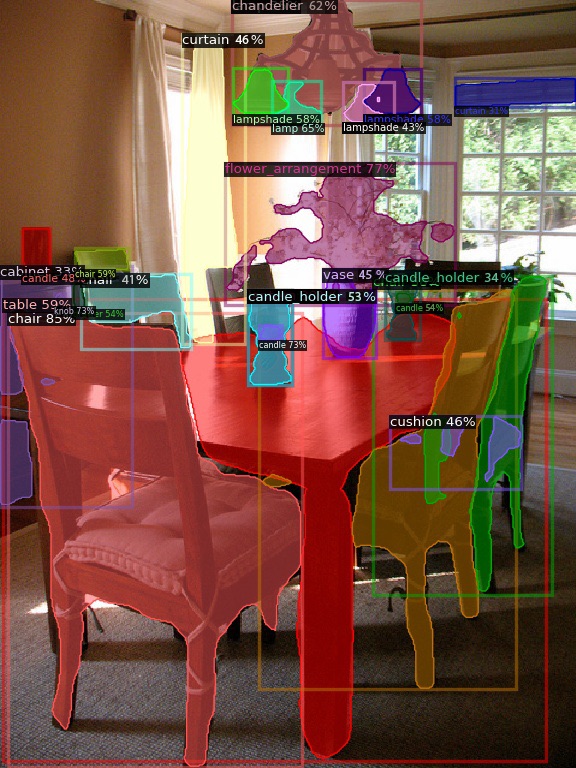}&
\includegraphics[width=0.32\linewidth]{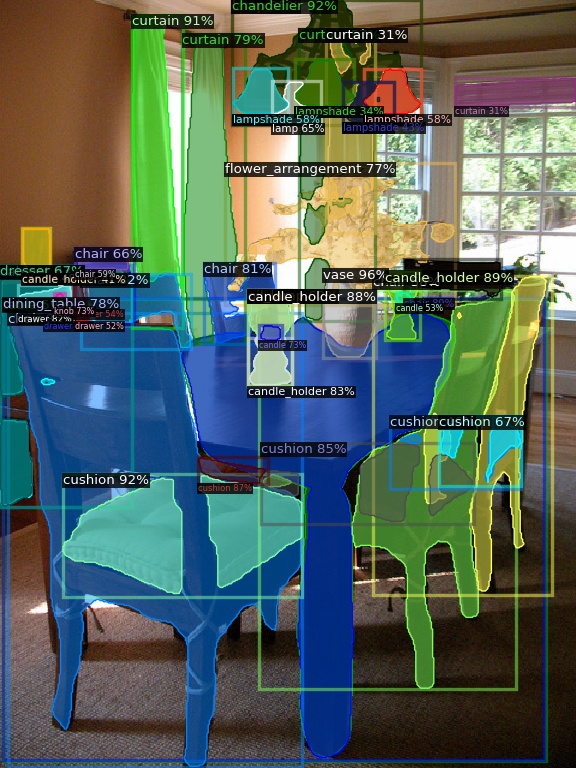}\\
Original & Our pre-trained teacher model  & Our semi-supervised student model\\
\end{tabular}
\caption{Qualitative example on the official LVIS val set.
}
\label{fig:original}  
\end{figure}
 
\fi 

\end{document}